\documentclass{article}
\usepackage{ijcai25}

\usepackage{times}
\usepackage{soul}
\usepackage{url}
\usepackage[hidelinks]{hyperref}
\usepackage[utf8]{inputenc}
\usepackage[small]{caption}
\usepackage{graphicx}
\usepackage{amsmath}
\usepackage{amsthm}
\usepackage{amssymb}
\usepackage{booktabs}
\usepackage{algorithm}
\usepackage{pifont}
\usepackage{algorithmic}
\usepackage{multirow}
\usepackage[switch]{lineno}

\title{HSRMamba: Contextual Spatial-Spectral State Space Model for Single Hyperspectral Image Super-Resolution}

\author{
Shi Chen$^1$\and
Lefei Zhang$^{1}$\footnote{Corresponding author}\And
Liangpei Zhang $^2$\\
\affiliations
$^1$National Engineering Research Center for Multimedia Software, School of Computer Science, Wuhan University, Wuhan, 430072, P. R. China\\
$^2$Aerospace Information Research Institute, Henan Academy of Sciences\\
\emails
\{chenshi@whu.edu.cn\}
}

\begin{document}

\maketitle

\begin{abstract}


Mamba has demonstrated exceptional performance in visual tasks due to its powerful global modeling capabilities and linear computational complexity, offering considerable potential in hyperspectral image super-resolution (HSISR). However, in HSISR, Mamba faces challenges as transforming images into 1D sequences neglects the spatial-spectral structural relationships between locally adjacent pixels, and its performance is highly sensitive to input order, which affects the restoration of both spatial and spectral details. In this paper, we propose HSRMamba, a contextual spatial-spectral modeling state space model for HSISR, to address these issues both locally and globally. Specifically, a local spatial-spectral partitioning mechanism is designed to establish patch-wise causal relationships among adjacent pixels in 3D features, mitigating the local forgetting issue. Furthermore, a global spectral reordering strategy based on spectral similarity is employed to enhance the causal representation of similar pixels across both spatial and spectral dimensions. Finally, experimental results demonstrate our HSRMamba outperforms the state-of-the-art  methods in quantitative quality and visual results. Code is available at: \url{https://github.com/Tomchenshi/HSRMamba}.
\end{abstract}

\section{Introduction}
Hyperspectral images (HSIs) typically comprise tens to hundreds of closely contiguous spectral bands spanning a broad spectral range, thereby capturing rich spectral and spatial information simultaneously \cite{XW2023}. This capability enables a more accurate characterization of intrinsic spectral properties and subtle variations in materials, and consequently facilitates widespread applications in fields such as agriculture \cite{LD2020}, medical diagnosis \cite{LF2014}, and remote sensing \cite{DD2023}. However, achieving high spectral resolution often necessitates sacrifices in spatial resolution due to the inherent constraints of imaging hardware and procedures. Hyperspectral image super-resolution (HSISR) addresses the challenge by converting low-resolution (LR) HSIs into the high-resolution (HR) HSIs, thereby enhancing spatial detail while preserving rich spectral information. 

\begin{figure}[t]
\centering
\includegraphics[width=0.48\textwidth]{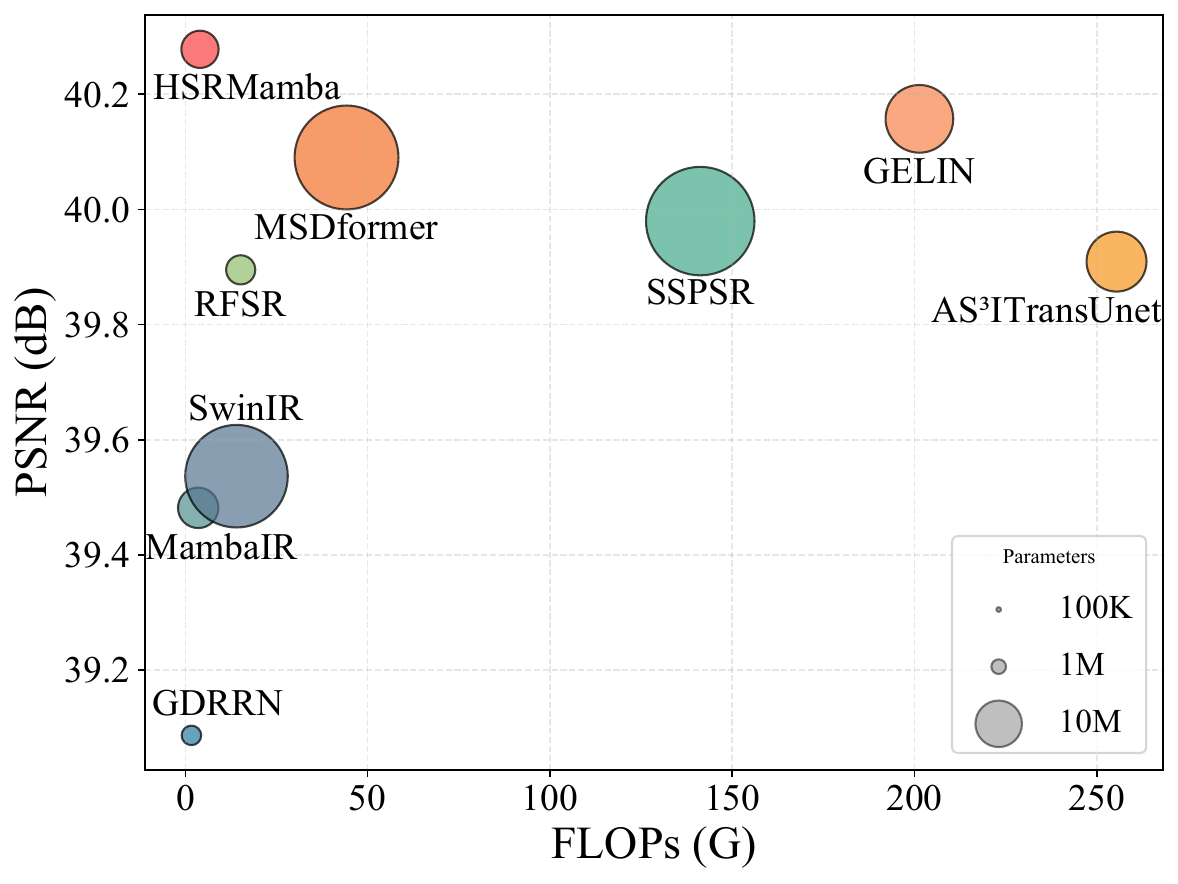}
\caption{Comparison of trade-offs between model performance and effectiveness on Chikusei at scale factor $\times4$. Our method achieves superior performance with relatively low computational cost.}
\label{complexity}
\end{figure}

\begin{figure*}[t]
\centering
\includegraphics[width=\textwidth]{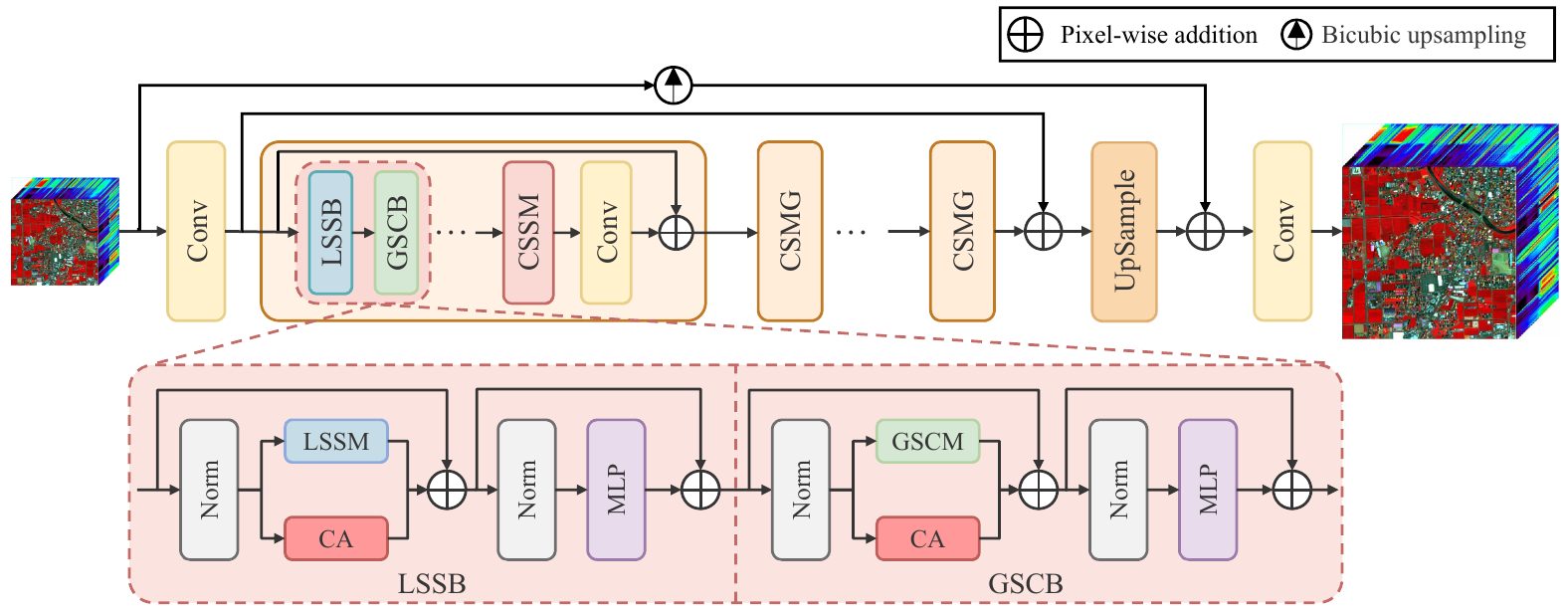}
\caption{The overview of the proposed HSRMamba. HSRMamba primarily comprises multiple Contextual Spatial-Spectral Mamba Groups (CSMGs). Each CSMG consists of several consecutive Contextual Spatial-Spectral Mamba Blocks, which are further composed of a Local Spatial-Spectral Mamba Module (LSSM) and a Global Spectral Correlation Mamba Module (GSCM).}
\label{HSRMamba}
\end{figure*}

Generally, HSISR can be categorized into fusion-based approaches \cite{GX2023,HY2023} and single hyperspectral image super-resolution (SHSR) \cite{ZZ2023}, depending on whether auxiliary information is employed. Although fusion-based strategies  can deliver superior results when aided by precisely registered additional images like multispectral image and panchromatic image, acquiring auxiliary information is often a significant challenge in real-world scenarios. Consequently, SHSR has attracted increasing interest. Over the past decade, deep learning SHSR methods have demonstrated remarkable advantages over traditional priors \cite{CW2023}, owing to their potent capacity for modeling complex nonlinear relationships. Among these methods, Transformer-based networks \cite{Interactformer,msdformer} have substantially enhanced HSISR performance by capturing long-range dependencies in spatial and spectral dimensions, thereby highlighting the critical importance of modeling long-range dependencies in HSISR. 
However, these models often face a trade-off between the global receptive field and computational efficiency \cite{LS2023,ZM2024,DW2024} due to the attention mechanism in Transformers, which severely limits further advancements in HSISR.

Recently, Mamba, an emerging state space model (SSM), has garnered significant attention for its ability to capture long-range dependencies with linear computational complexity. Despite its theoretical potential to address the challenge in attention mechanism, Mamba suffers from some issues for HSISR. Concretely, When transforming hyperspectral images into 1D sequences, the Mamba model overlooks the spatial-spectral structure among locally adjacent pixels, severely limiting its performance. Moreover, the output of Mamba is highly dependent on the input order, and a simple unfolding process disregards the modeling of spatial-spectral relationships between highly similar pixels, which is crucial for hyperspectral image restoration tasks.


To address these challenges, we propose HSRMamba, a state space model designed to efficiently capture both local and global long-range spatial-spectral dependencies. To the best of our knowledge, HSRMamba is the first attempt to employ Mamba tailored for SHSR. Specifically, we design a local spatial-spectral partitioning (LSSP) mechanism, which divides the 3D features into local spatial and spectral windows. By introducing the bidirectional SSM \cite{ssumamba} (BSSM), a local spaital-spectral Mamba module (LSSM) is constructed to establish causal relationships among neighboring pixels within these local 3-D windows, which enhances the network’s local modeling capacity. Furthermore, we devise a global spectral reordering mechanism (GSRM), which rearranges the global spectra based on global spectral similarity. Subsequently we employ the Global Spectral Correlation Mamba module (GSCM) to strengthen causal modeling between highly similar pixels, facilitating detail reconstruction in hyperspectral images. This process further strengthens causal relationships among pixels in the spatial and spectral dimensions that share high similarity. In summary, the main contributions of our work are as follows:

\begin{itemize}

\item We propose HSRMamba, the first SSM tailored for SHSR, which efficiently establishes local and global long-range spatial-spectral causal relationships.

\item We devise a local spatial-spectral partitionin mechanism that captures patch-wise long-range spatial-spectral dependencies within a 3D window, thereby alleviating the inherent local pixel forgetting issue.

\item We develop a global spectral-correlation Mamba module that globally extracts long-range spatial-spectral features by reordering spectra based on global spectral similarity, thus bolstering causal modeling among highly similar pixels.

\item Extensive experiments on various datasets demonstrates the superiority and effectiveness of our proposed technique over the state-of-the-art methods.

\end{itemize}

\section{Related Works}

\subsection{Single Hyperspectral Image Super-resolution}
Without additional auxiliary data (e.g., panchromatic or multispectral images), SHSR exhibits broader applicability compared to fusion-based methods \cite{DL2019,VG2022,DQ2023}. SHSR can be categorized into traditional methods \cite{DF2017} based on handcrafted priors and deep learning-based approaches \cite{WH2023}. Over the past decade, numerous learning-based approaches, such as 3D convolution-based methods \cite{3dfcnn,MCNet,FL2021,ERCSR}, group strategy-based methods \cite{GDRRN,SSPSR,dualsr,RFSR,GELIN}, and Transformer-based methods \cite{3DTHSR,CST}, have demonstrated significantly superior performance compared to traditional approaches. 


In the above methods, convolution-based with 2D or 3D convolutions mainly focused on local spatial-spectral features, overlooking long-range spatial-spectral dependencies. Recently, Cai et al. \cite{CL2022} proposed a spectral-wise multi-head self-attention for HSI reconstruction. Wang et al. \cite{3DTHSR} introduced 3D-THSR, combining spectral self-attention with 3D convolutions to model spatial-spectral features in a global receptive field. Hu et al. \cite{HW2024} transfered the HSI SR to the abundance domain with spectral-wise non-local attention to effectively incorporates valuable knowledge. Nevertheless, the computational complexity of Transformer networks grows quadratically with the input size, which significantly increases the demand for hardware resources when dealing with high-dimensional data such as hyperspectral images. While window-based self-attention mechanisms reduce computational costs by limiting the attention range, they fail to fully address the issue of high complexity. To overcome this limitation, we propose the Mamba network, a linear modeling framework designed to efficiently achieve hyperspectral image super-resolution reconstruction.

\subsection{State Space Model}
State Space Model \cite{SSM} are a mathematical framework designed to model temporal or sequential dependencies efficiently. More recently, Mamba, a SSMs-based model with linear computational complexity, has garnered significant attention for outperforming Transformers in natural language processing \cite{mamba} and computer vision \cite{visionmamba,localmamba} tasks. Subsequently, some Mamba networks designed for low-level vision tasks \cite{mambair,HImamba,fmsr,MambaFormerSR,ssumamba} have been proposed. Nonetheless, these methods are not suited for hyperspectral images. Firstly, they fail to consider the rich spectral information and the correlations between spatial and spectral dimensions inherent in hyperspectral data. Secondly, the Mamba network suffers from a local pixel forgetting issue, limiting its ability to model the structural information across spatial and spectral dimensions. Lastly, the performance of the Mamba network is highly dependent on the input sequence order, making it unable to fully exploit the relationships between highly similar pixels, which are crucial for image restoration.

\section{Method}

\subsection{Preliminaries: State Space Models}
SSMs provide a mathematical framework for modeling systems governed by latent states and their transitions over time. SSMs effectively model temporal dependencies in sequential data, and can be formulated as linear ordinary differential equation:
\begin{equation}
\label{eq1}
\begin{aligned}
h'(t) &= \mathbf{A} h(t) + \mathbf{B} x(t), \\
y(t) &= \mathbf{C} h(t) + \mathbf{D} x(t),
\end{aligned}
\end{equation}
where $h(t) \in \mathbb{R}^N$ is the hidden state vector at time \(t\), $\mathbf{A} \in \mathbb{R}^{N \times N}$, $\mathbf{B} \in \mathbb{R}^{N \times 1}$, $\mathbf{C} \in \mathbb{R}^{1 \times N}$ and $\mathbf{D} \in \mathbb{R}$ are the weight parameters, and $N$ is the hidden state size.

Then, Eq.\ref{eq1} can be discretized using the zeroth-order hold (ZOH) rule, which transforms the continuous parameters $\mathbf{A}$, $\mathbf{B}$ to discrete parameters $\overline{\mathbf{A}}$, $\overline{\mathbf{B}}$ by the the timescale parameter $\Delta$. It can be defined as:
\begin{equation}
\begin{aligned}
\overline{\mathbf{A}} &= \exp(\Delta \mathbf{A}), \\
\overline{\mathbf{B}} &= (\Delta \mathbf{A})^{-1} \left( \exp(\Delta \mathbf{A}) - \mathbf{I} \right) \cdot \Delta \mathbf{B}.
\end{aligned}
\end{equation}

After discretization, the discrete version of Eq. \ref{eq1} can be defined in the following RNN form:
\begin{equation}
\label{eq3}
\begin{aligned}
h_k &= \overline{\mathbf{A}} h_{k-1} + \overline{\mathbf{B}} x_k, \\
y_k &= \mathbf{C} h_k + \mathbf{D} x_k.
\end{aligned}
\end{equation}
Furthermore, the SSM computation can be extended to a convolutional form as:
\begin{equation}
\begin{aligned}
\overline{\mathbf{K}} &\triangleq \left(\mathbf{C}\overline{\mathbf{B}}, \mathbf{C}\overline{\mathbf{A}}\overline{\mathbf{B}}, \dots, \mathbf{C}\overline{\mathbf{A}}^{L-1}\overline{\mathbf{B}}\right), \\
\mathbf{y} &= \mathbf{x} \ast \overline{\mathbf{K}},
\end{aligned}
\end{equation}
where $L$ is the length of the input sequence and $\ast$ is the convolution operation, and $\overline{\mathbf{K}} \in \mathbb{R}^{L}$ respresents a structured convolutional kernel. Recently, leveraging a dynamic representation mechanism, Selective State Space Model (Mamba) enhances the long-range modeling capabilities of state space models with linear computational complexity.

\subsection{Overview Architecture}

\begin{figure*}[th]
\centering
\includegraphics[width=0.92\textwidth]{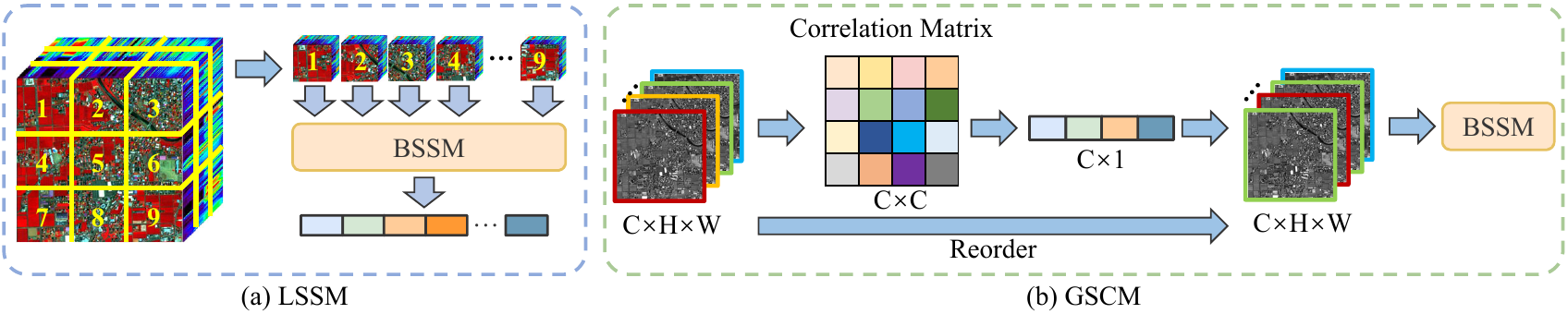}
\caption{The pipeline of the designed LSSM (left) and GSCM (right).}
\label{pipeline}
\end{figure*}

As shown in Fig. \ref{HSRMamba}, the overall architecture of HSRMamba consists of three primary components: a shallow feature extraction module, a deep feature extraction module, and an upsampling module. Assuming the input LR hyperspectral image is \(I_{\rm LR} \in \mathbb{R}^{H \times W \times B}\), where \(H\), \(W\), and \(B\) represent the spatial height, width, and the number of spectral bands, respectively. The SR hyperspectral image is denoted as \(I_{\rm SR} \in \mathbb{R}^{sH \times sW \times B}\), where \(s\) is the scaling factor for super-resolution reconstruction. The shallow features is extracted by using $3\times3$ convolutional layer. This process captures the structural information and adjusts the feature channel dimensions, which can be denoted as:
\begin{equation}
    F_{0} = f_{\rm s}(I_{\rm LR}),
\end{equation}
where $f_{\rm s}(\cdot)$ represents $3\times3$ convolution, and $F_{0}$ denotes the extracted shallow features. Next, the shallow features are passed through the deep feature extraction module, consisting of multiple cascaded contextual spatial-spectral Mamba group (CSMG). The long-range spatial-spectral dependencies in deep features are extracted locally and globally. The process is defined as:
\begin{equation}
    F_{n} = H_{\rm n}(F_{n-1}), \quad n = 1,2,\dots,N,
\end{equation}
where $H_{\rm n}(\cdot)$ denotes the function of the \(n\)-th CSMG, and $F_{n}$ represents the deep spatial-spectral features extracted by the $n$-th CSMG. Each CSMG consists of multiple consecutive contextual spatial-spectral Mamba modules (CSSM), each of which is composed of a local spatial-spectral Mamba block (LSSB) and a global spectral-correlation Mamba block (GSCB). Additionally, convolution is employed at the tail of each group to introduce inductive bias into the network. 

By leveraging long skip connections between shallow and deep features, the network focuses on learning high-frequency information beneficial for image reconstruction, improving both training efficiency and reconstruction performance. The merged deep features are then passed through a pixelshuffle layer to produce the upsampling deep features. Finally, SR HSI is obtained through long skip connections between the input features after bicubic upsampling and deep features.

\subsection{Local Spaital-Spectral Mamba Module}
As the performance of Mamba is heavily reliant on the order of the input, converting a hyperspectral image into a 1-D sequence for sequential scanning poses significant challenges in effectively establishing causal relationships between locally adjacent pixels across spatial and spectral dimensions. This inherent limitation greatly hinders the Mamba network's effectiveness in HSISR. Therefore, we design a local scanning mechanism capable of precisely capturing detailed connections between locally adjacent pixels, effectively addressing the limitations of Mamba in local causal modeling. 

As shown in Fig.~\ref{HSRMamba}, LSSB consists of layer normalization, LSSM, channel atteniton (CA), and linear mapping layer. LSSM extract long-range spatial-spectral information from locally scanned regions. Additionally, CA mechanism is  introduced to add inductive bias property into the network. Finally, the processed information is refined through linear layers, enhancing the module's capability for deep feature extraction. Assuming the input of the $j$-th local spatial-spectral Mamba module in the contextual spatial-spectral Mamba component is \(F_{j}\), the above process can be expressed as:
\begin{equation}
\begin{aligned}
\hat{F}_{L}^{j} &= {\rm LSSM}({\rm LN}(F_{j})) + {\rm CA}({\rm LN}(F_{j})) + F_{j}, \\
F_{L}^{j} &= {\rm MLP}({\rm LN}(\hat{F}_{L}^{j})) + \hat{F}_{L}^{j},
\end{aligned}
\end{equation}
where \({\rm LN}(\cdot)\) denotes the layer normalization function, \({\rm LSSM}(\cdot)\) represents the local spatial-spectral Mamba block, \({\rm CA}(\cdot)\) denotes the channel attention mechanism, \({\rm MLP}(\cdot)\) represents the linear layer function, and \(F_{L}^{j}\) is the final output of the module. LSSM consists of a LSSP and a BSSM \cite{ssumamba}. As illustrated in Fig.~\ref{pipeline}, the spatial-spectral partitioning mechanism divides the input features into \(\frac{H}{h} \times \frac{W}{w} \times \frac{C}{c}\) local 3D feature blocks, each of size \(h \times w \times c\), along the spatial and spectral dimensions. 
BSSM captures long-range spatial-spectral dependencies within each 3D local feature block. This process effectively addresses the local pixel forgetting issue inherent in traditional Mamba modules.

\begin{figure}[t]
\centering
\includegraphics[width=0.42\textwidth]{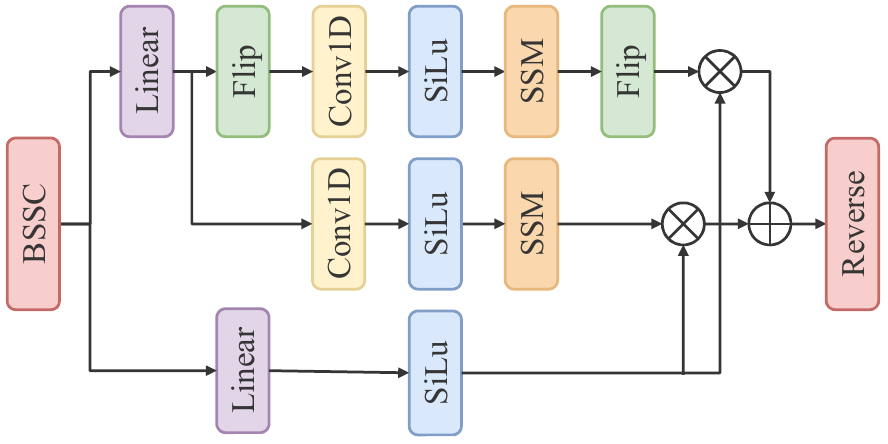}
\caption{The pipeline of the proposed BSSM.}
\label{BSSM}
\end{figure}

In this paper, we introduce a BSSM \cite{ssumamba} as the foundational unit for global spatial-spectral feature extraction. As shown in Fig.~\ref{BSSM}, the BSSM module takes the feature sequence from LSSP or GSRM as input of bidirectional branches and models comprehensive global spatial-spectral causal relationships from different directions via bidirectional spatial-spectral continuous scanning (BSSC). This design complementarily enhances contextual long-range dependency modeling, making it suited for hyperspectral images. 

\begin{table*}[t]
\centering
\setlength{\tabcolsep}{0.0035\hsize}{
\begin{tabular}{ccccccc|ccccc}
\toprule[1pt]
\multirow{2}{*}{Method} & \multirow{2}{*}{Scale} & \multicolumn{5}{c|}{Chikusei} & \multicolumn{5}{c}{Houston} \\
 &  & \multicolumn{1}{c}{PSNR$\uparrow$} & \multicolumn{1}{c}{SSIM$\uparrow$} & \multicolumn{1}{c}{SAM$\downarrow$} & \multicolumn{1}{c}{CC$\uparrow$} & \multicolumn{1}{c|}{ERGAS$\downarrow$} & \multicolumn{1}{c}{PSNR$\uparrow$} & \multicolumn{1}{c}{SSIM$\uparrow$} & \multicolumn{1}{c}{SAM$\downarrow$} & \multicolumn{1}{c}{CC$\uparrow$}  & \multicolumn{1}{c}{ERGAS$\downarrow$} \\
\midrule
GDRRN \cite{GDRRN} & $\times4$ & 39.0864 & 0.9265 & 3.0536 & 0.9421 & 5.7972 & 44.2964 & 0.9730 & 2.5347 & 0.9760 & 2.4700 \\
SwinIR \cite{SWINIR} & $\times4$ & 39.5366 & 0.9364 & 2.8327 & 0.9456 & 5.6280 & 46.0971 & \underline{0.9808} & 1.9463 & 0.9864 & 2.0039 \\
MambaIR \cite{mambair} & $\times4$ & 39.4816 & 0.9353 & 2.9178 & 0.9456 & 5.6303 & 45.5947 & 0.9786 & 1.9953 & 0.9849 & 2.1239 \\
SSPSR \cite{SSPSR} & $\times4$ & 39.9797 & 0.9393 & 2.4864 & 0.9528 & 5.1905 & 45.6017 & 0.9778 & 1.9650 & 0.9850 & 2.1380 \\
RFSR \cite{RFSR} & $\times4$ & 39.8950 & 0.9382 & 2.4656 & 0.9517 & 5.2334 & 45.8677 & 0.9792 & 1.8304 & 0.9858 & 2.0659 \\
GELIN \cite{GELIN} & $\times4$ & \underline{40.1573} & \underline{0.9410} & 2.4266 & 0.9543 & \underline{5.0314} & 45.8715 & 0.9790 & 1.8759 & 0.9859 & 2.0778 \\
AS\textsuperscript{3}ITransUNet \cite{AS3Net} & $\times4$ & 39.9093 & 0.9377 & 2.6056 & 0.9519 & 5.1900 & 45.8819 & 0.9792 & 1.8679 & 0.9862 & 2.0731 \\
MSDformer \cite{msdformer} & $\times4$ & 40.0902 & 0.9405 & \underline{2.3981} & \underline{0.9539} & 5.0818 & \underline{46.2015} & 0.9807 & \underline{1.7776} & \underline{0.9870} & \underline{1.9962} \\
Ours & $\times4$ & \textbf{40.2781} & \textbf{0.9441} & \textbf{2.3160} & \textbf{0.9557} & \textbf{5.0131} & \textbf{46.9653} & \textbf{0.9838} & \textbf{1.6577} & \textbf{0.9891} & \textbf{1.8277} \\
\midrule
GDRRN \cite{GDRRN} & $\times8$ & 34.7395 & 0.8199 & 5.0967 & 0.8381 & 9.6464 & 38.2592 & 0.9085 & 4.9045 & 0.9138 & 4.9135 \\
SwinIR \cite{SWINIR} & $\times8$ & 34.8785 & 0.8307 & 5.0413 & 0.8465 & 9.4743 & 39.4013 & 0.9194 & 4.0586 & 0.9370 & 4.3333 \\
MambaIR \cite{mambair} & $\times8$ & 35.1962 & 0.8365 & 4.7499 & 0.8556 & 9.0543 & 39.3049 & 0.9193 & 4.1935 & 0.9358 & 4.3648 \\
SSPSR \cite{SSPSR} & $\times8$ & 35.1643 & 0.8299 & 4.6911 & 0.8560 & 9.0504 & 39.2844 & 0.9164 & 4.2673 & 0.9346 & 4.4212 \\
RFSR \cite{RFSR} & $\times8$ & 35.5049 & 0.8405 & 4.2785 & 0.8661 & 8.6338 & 39.4899 & 0.9211 & 3.8403 & 0.9379 & 4.2967 \\
GELIN \cite{GELIN} & $\times8$ & \underline{35.6496} & \underline{0.8464} & \underline{4.1354} & \underline{0.8707} & \underline{8.4520} & 39.6387 & 0.9216 & 3.9231 & 0.9393 & 4.2453 \\
AS\textsuperscript{3}ITransUNet \cite{AS3Net} & $\times8$ & 35.4999 & 0.8408 & 4.4746 & 0.8661 & 8.6793 & \underline{39.8196} & \underline{0.9254} & 3.9035 & \underline{0.9422} & \underline{4.1466} \\
MSDformer \cite{msdformer} & $\times8$ & 35.5914 & 0.8452 & 4.1381 & 0.8693 & 8.5203 & 39.7452 & 0.9227 & \underline{3.6613} & 0.9417 & 4.2110 \\
Ours & $\times8$ & \textbf{35.6812} & \textbf{0.8474} & \textbf{4.1148} & \textbf{0.8724} & \textbf{8.4508} & \textbf{39.9309} & \textbf{0.9265} & \textbf{3.5627} & \textbf{0.9443} & \textbf{4.1138}\\
\bottomrule[1pt]
\end{tabular}
}
\caption{Quantitative performance on the Chikusei dataset and Houston dataset at different scale factors. Bold represents the best result and underline represents the second best.}
\label{tab1}
\end{table*}

\subsection{Global Spectral Correlation Mamba Module}

In the Mamba network, the output of the current input relies solely on the preceding input in the data sequence. However, the recovery of fine details during the super-resolution process heavily depends on global similar pixels. To address this limitation, we propose a global spectral reordering mechanism and design the global spectral correlation Mamba.

As shown in Fig.~\ref{HSRMamba}, GSCB consists of layer normalization, the designed GSCM module, CA, and linear mapping layer. GSCM enhances glboal spatial-spectral causal modeling through the proposed global spectral reordering mechanism. Additionally, this module incorporates a channel attention mechanism to introduce inductive bias into the network. Finally, a linear layer is employed to enhance the representation capabilities of this paper.

Assuming the input to the $j$-th global spectral correlation Mamba module in the contextual spatial-spectral Mamba component is \(F_{j}\), the above process can be expressed as:
\begin{equation}
\begin{aligned}
\hat{F}_{G}^{j} &= {\rm GSCM}({\rm LN}(F_{L})) + {\rm CA}({\rm LN}(F_{L})) + F_{L}, \\
F_{G}^{j} &= {\rm MLP}({\rm LN}(\hat{F}_{G}^{j})) + \hat{F}_{G}^{j},
\end{aligned}
\end{equation}
where \({\rm LN}(\cdot)\) denotes the layer normalization function, \({\rm GSCM}(\cdot)\) represents the global spectral correlation Mamba block, \({\rm CA}(\cdot)\) denotes the channel attention mechanism, and \({\rm MLP}(\cdot)\) represents the linear layer function. \(\hat{F}_{G}^{j}\) is the intermediate output of the $j$-th global spectral correlation Mamba module, and \(F_{G}^{j}\) is the final output of the module.

As illustrated in Fig.~\ref{pipeline}, the global spectral correlation Mamba consists of the GSRM and the BSSM. The global spectral reordering mechanism first computes the correlation coefficient matrix between spectral features. It then calculates the average of correlation coefficients among each spectrum as the global correlation value. Finally, the module reorders the spectral features based on their global correlation values, ensuring that pixels with high spectral correlation are closer in the spatial-spectral dimensions. This process significantly enhances the Mamba module's performance in extracting long-range spatial-spectral features.

\subsection{Loss Function}
The network is optimized using three losses: $l_1$ loss, spectral angle mapper (SAM) loss, and gradient loss in both spatial and spectral domains. The $l_1$ loss calculates the absolute pixel-wise difference between the reconstructed and original hyperspectral images, encouraging sharper and more detailed results compared to $l_2$ loss. The SAM loss ensures spectral consistency by considering the spectral characteristics of the data. Gradient loss enhances image sharpness by focusing on differences between adjacent pixels. The total loss is formulated as:
\begin{equation}
\mathcal{L}_{total}(\theta) = \mathcal{L}_1 + \lambda_s \mathcal{L}_{sam} + \lambda_g \mathcal{L}_{gra},
\end{equation}
where $\lambda_s$ and $\lambda_g$ are hyper-parameters that balance the losses, with $\lambda_s = 0.3$ and $\lambda_g = 0.1$ used empirically. The detailed function can be expressed as:
\begin{equation}
\begin{aligned}
\mathcal{L}_1(\theta) &= \frac{1}{N} \sum\limits_{n=1}^{N} \left\| H_{hr}^n - H_{sr}^n \right\|_1,\\
\mathcal{L}_{sam}(\theta) &= \frac{1}{N} \sum\limits_{n=1}^{N} \frac{1}{\pi} \arccos\left( \frac{H_{hr}^n \cdot H_{sr}^n} {\left\| H_{hr}^n \right\|_2 \cdot \left\| H_{sr}^n \right\|_2} \right),\\
\mathcal{L}_{gra}(\theta) &= \frac{1}{N} \sum\limits_{n=1}^{N} \left\| M(H_{hr}^n) - M(H_{sr}^n) \right\|_1, \\
\end{aligned}
\end{equation}
where $N$ is the batch size, $H_{hr}^n$ and $H_{sr}^n$ represent the $n$-th HR and SR hyperspectral images, and $M(\cdot)$ denotes gradients in the horizontal, vertical, and spectral directions.

\section{Experiments}
\begin{figure*}[t]
\centering
\includegraphics[width=\textwidth]{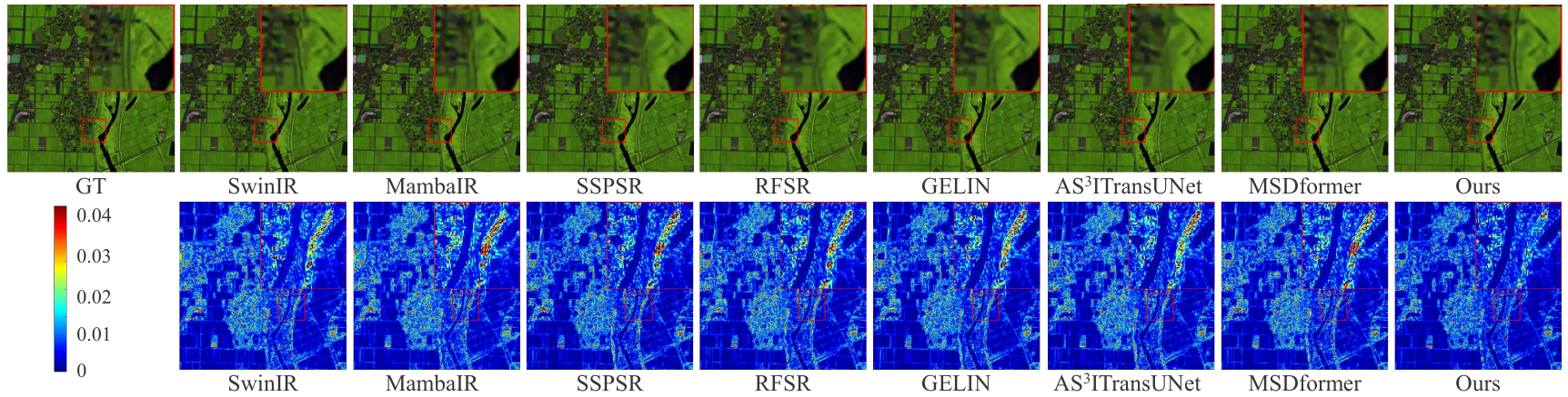}
\caption{Visual results on the Chikusei dataset with spectral bands 70-100-36 as R-G-B at scale factor $\times$4.}
\label{chi}
\end{figure*}

\begin{figure*}[t]
\centering
\includegraphics[width=\textwidth]{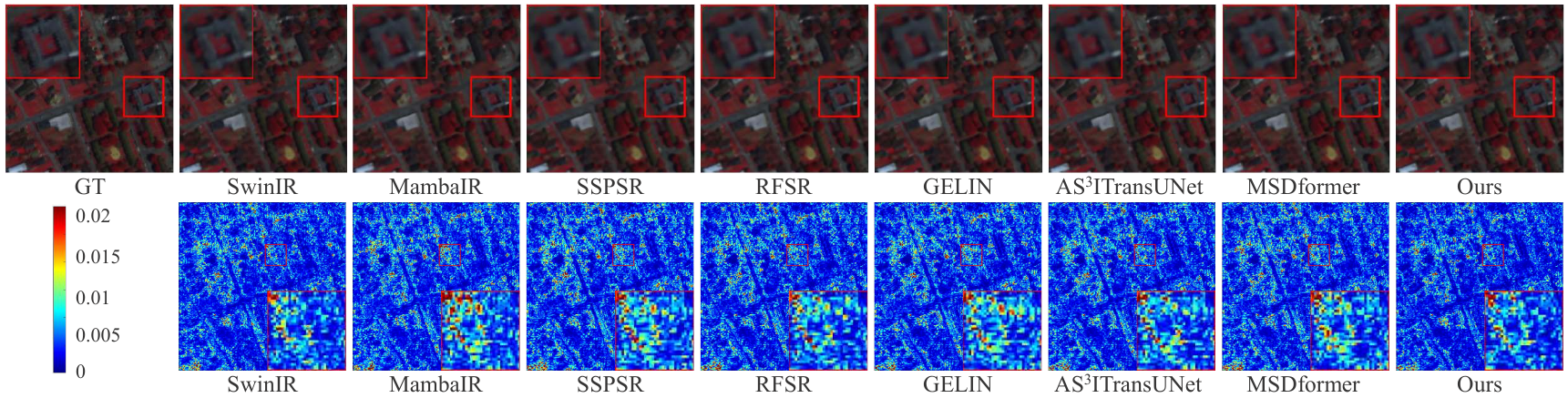}
\caption{Visual results on the Houston dataset with spectral bands 26-20-10 as R-G-B at scale factor $\times$4.}
\label{hou}
\end{figure*}

\subsection{Experimental Settings}
\subsubsection{Datesets}
We conducted experiments on three hyperspectral image datasets: Chikusei \cite{chikusei}, Houston2018, and Pavia Center \cite{pavia}. The experiments on Pavia Center are presented in the supplementary materials. The Chikusei dataset, captured with the Headwall Hyperspec-VNIR-C sensor, includes 128 spectral bands over agricultural and urban areas in Japan, with a spatial resolution of $2517 \times 2335$ pixels. The Houston 2018 dataset is a part of the 2018 IEEE
GRSS Data Fusion Contest, acquired by the ITRES CASI 1500 imager, covers the University of Houston and surrounding urban areas with 48 spectral bands and a resolution of $4172 \times 1202$ pixels. 

\subsubsection{Compared Methods and Metrics}
We compare the proposed method with 8 deep learning approaches, including Transformer-based methods SwinIR \cite{SWINIR} for natural images, Mamba-based methods MambaIR \cite{mambair} for natural images, and hyperspectral image group-based methods such as GDRRN \cite{GDRRN}, SSPSR \cite{SSPSR}, RFSR \cite{RFSR}, and GELIN \cite{GELIN}. We also include Transformer-based SHSR methods like AS\textsuperscript{3}ITransUNet and MSDformer \cite{msdformer}. The performance of these methods is evaluated using six commonly used metrics in both spatial and spectral dimensions, including peak signal-to-noise ratio (PSNR), structure similarity (SSIM), spectral angle mapper (SAM), cross correlation (CC), root-mean-squared error (RMSE), and erreur relative global adimensionnellede synthese (ERGAS).

\subsubsection{Implementation Details}
The kernel size of the convolution is set to $3 \times3$. We set the number of channels $C$ to 64, the number of CSMG to 4, and the number of CSSM to 2. The initial learning rate is $1e^{-4}$, halving every 100 epochs until reaching 400 epochs. Following \cite{RCAN}, the reduction ratio in channel attention (CA) is set to 16. During training, the Adam optimizer with Xavier initialization is used with a mini-batch size of 8. For image reconstruction, we use a progressive upsampling strategy via PixelShuffle \cite{SC2016} to reduce parameters. The model is implemented in Pytorch and trained on NVIDIA RTX 4090 GPUs.


\subsection{Comparison Results}

\subsubsection{Experiments on the Chikusei Dataset}
For Chikusei dataset, 4 non-overlapping images with the size of $512\times 512\times 128$ are cropped from the top region. The remaining area is cropped into overlapping HR images for training (10\% randomly selected for validation). The spatial size of the LR for training  is $32\times32$, while the corresponding HR sizes at scale factors $\times4$, and $\times8$ are $128\times128$, and $256\times256$, respectively. All LR patches are generated by Bicubic downsampling at different scales.

Table~\ref{tab1} presents the quantitative results of our method and the compared approaches at different scale factors on the Chikusei dataset. The best results are highlighted in bold, while the second-best results are underlined. Our method surpasses SSPSR by 0.29 dB in PSNR and 0.32 in SAM at $\times4$, as group-based methods like SSPSR fail to leverage global spatial and spectral information effectively. Methods for natural images like SwinIR and MambaIR fail to fully utilize the rich spectral information and spatial-spectral correlations in hyperspectral images, resulting in poor spectral performance. Transformer-based SHSR methods like MSDformer and AS\textsuperscript{3}ITransUNet achieve better SR performance than the above methods by modeling long-range dependencies. Notably, HSRMamba outperforms other methods for all metrics at scale factors $\times4$, and $\times8$, demonstrating the superiority and effectiveness of our method.

\begin{figure}[t]
\centering
\includegraphics[width=0.4\textwidth]{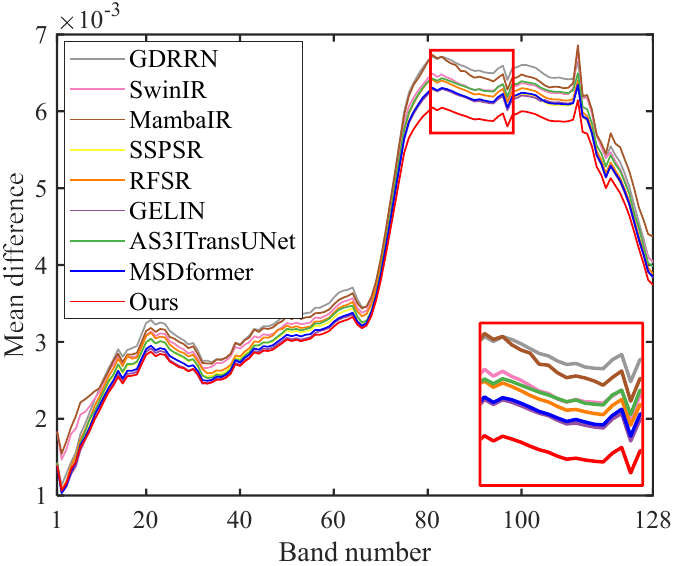}
\caption{Mean spectral difference curves of different methods on the Chikusei dataset at scale factors $\times$4.}
\label{chi_mean}
\end{figure}

As shown in Fig.~\ref{chi}, group-based methods like GDRRN, often introduce blur details due to the limited spatial-spectral modeling. Transformer-based methods, such as MSDformer, consider global spatial and spectral information simultaneously, resulting in clearer boundaries and fewer artifacts compared to methods designed for natural images, such as MambaIR. HSRMamba reconstructs HR hyperspectral images with clearer and sharper details, which demonstrates that our HSRMamba can efficiently model long-range spatial-spectral dependencies. Additionally, mean error maps across all spectra in Fig.~\ref{chi} show the reconstruction accuracy for individual pixels, with bluer regions indicating higher accuracy. And the mean spectral difference curves at $\times 4$ in Fig.~\ref{chi_mean} evaluate the SR results from a spectral perspective. From the above visual results, it is evident that our method achieves results closer to the ground truth in both spatial and spectral dimensions compared to other methods.

\subsubsection{Experiments on the Houston 2018 Dataset}
Similar to Chikusei dataset, 8 images from the Houston2018 dataset with the size of $256\times 256\times 48$ are cropped from the top region for testing. The spatial resolution of LR and HR training patches is consistent with the Chikusei dataset.

The quantitative results of all methods on Houston dataset are shown in Table~\ref{tab1}. HSRMamba outperforms the compared methods for all metrics at different scale factors. The visual results and mean error maps of all algorithms are presented in Fig.~\ref{hou}. We can also observe that HSRMamba provides more accurate results compared with other approaches.

\subsection{Ablation Study}
In this paper, we conduct the ablation experiments at factor scale $\times4$ on the Chikusei dataset. Additional ablation studies are listed in the supplementary materials.
\subsubsection{Effects of LSSP and GSRM} 
\begin{table}[t]
\centering
\begin{tabular}{ccccc}
\toprule[1pt]
LSSP & GSRM & PSNR$\uparrow$ & SSIM$\uparrow$   & SAM$\downarrow$  \\ 
\midrule
\ding{55}   & \ding{55}    & 40.0714 & 0.9420 & 2.4078 \\
\ding{51} &  \ding{55}    & 40.1532 & 0.9427 & 2.3812 \\
\ding{55} &  \ding{51}    & 40.1849 & 0.9430 & 2.3662 \\
\ding{51} &  \ding{51}  & \textbf{40.2781} & \textbf{0.9441} & \textbf{2.3160} \\
\bottomrule[1pt]
\end{tabular}
\caption{Quantitative performance of different components evaluated on Chikusei dataset at the scale factor $ \times 4$.}
\label{tab2}
\end{table}

LSSP effectively addresses the issue of local forgetting and GSRM relieves the sensitivity of the Mamba network to input order. As shown in Table~\ref{tab2}, when LSSP is removed, PSNR of the network decreases by 0.09 dB, demonstrating the effectiveness of LSSP. The PSNR of model without GSRM decreases by 0.12 dB compared to the original method, demonstrating the effectiveness of GSRM. Finally, when both LSSP and GSRM are removed, the performance drops significantly, further confirming the effectiveness of our method.

\subsubsection{Effects of the Number of Groups}
\begin{table}[t]
\centering
\setlength{\tabcolsep}{0.016\hsize}{
\begin{tabular}{ccccc}
\toprule[1pt]
Number ($N$) & Params.($\times 10^{6}$) & PSNR$\uparrow$    & SSIM$\uparrow$   & SAM$\downarrow$ \\
\midrule
$N=2$ &1.0807 & 40.1797 & 0.9428 & 2.3479  \\
$N=4$ &1.6073 &\textbf{40.2781} & \textbf{0.9441} & \textbf{2.3160}  \\
$N=6$ &2.1339 & 40.2312 & 0.9430 & 2.3523  \\
\bottomrule[1pt]
\end{tabular}
}
\caption{Quantitative comparisons of the number of Mamba groups over the Chikusei testing dataset at scale factor $\times$4.}
\label{tab3}
\end{table}

HSRMamba consists of multiple consecutive Mamba groups, and Table \ref{tab3} shows the impact of the number of Mamba groups $N$. When $N=2$, the performance is the weakest. As $N$ increases to 4, the quantitative metrics improve. However, setting $N=6$ leads to a decline in performance. This is mainly due to the increased network depth, causing the model to overfit. Therefore, considering both the experimental results and the model parameters, we set $N=4$ in the paper.

\subsubsection{Parameter and Complexity Analysis}
To evaluate the computational efficiency of the proposed HSRMamba, we compare the model parameters, FLOPs, and PSNR results for different methods. As shown in Fig.~\ref{complexity}, our method achieves better SR results with lower computational complexity and fewer parameters compared to other methods, demonstrating the effectiveness and efficiency of our approach. This indicates that our method strikes an excellent balance between model complexity and performance.

\section{Conclusion}
This paper presents HSRMamba, a contextual spatial-spectral relationship modeling algorithm designed for efficient HSISR. To address the issue of local pixel forgetting in hyperspectral images, we propose the LSSP to establish patch-wise long-range spatial-spectral correlations. Additionally, to overcome the challenge of insufficient causal modeling between highly similar pixels, we leverage the GSRM that rearranges the spectral dimension based on global spectral correlations. Using these algorithms, we construct the CSSM to efficiently capture long-range spatial-spectral dependencies in hyperspectral images. The CSSM module is composed of LSSM and GSCM, which enhance causal modeling by considering both local and global perspectives. Finally, extensive comparative experiments on different datasets and ablation studies validate the superiority and effectiveness of our proposed approach.


\bibliographystyle{named}
\bibliography{ijcai25}

\end{document}